\documentclass[manuscript,screen, nonacm]{acmart}

\setcopyright{none}
\renewcommand\footnotetextcopyrightpermission[1]{}

\usepackage{booktabs}
\usepackage{float}
\usepackage{longtable}
\usepackage{tikz}
\usepackage{pdfpages}
\usepackage{nicefrac}
\usepackage{microtype}
\usepackage{array}
\begin{document}

\title{Fairness in Explainable AI Narratives: Decomposing Social Disparities in LLM-Generated Explanations}

\author{Wannes Van den Bulck}
\affiliation{%
  \institution{University of Antwerp}
  \department{Department of Engineering Management}
  \city{Antwerp}
  \country{Belgium}}
\email{wannes.vandenbulck@uantwerpen.be}

\author{Sofie Goethals}
\affiliation{%
  \institution{University of Antwerp}
  \department{Department of Engineering Management}
  \city{Antwerp}
  \country{Belgium}}
\email{sofie.goethals@uantwerpen.be}

\author{David Martens}
\affiliation{%
  \institution{University of Antwerp}
  \department{Department of Engineering Management}
  \city{Antwerp}
  \country{Belgium}}
\email{david.martens@uantwerpen.be}

\authorsaddresses{Authors' Contact Information: Wannes Van den Bulck, Sofie Goethals and David Martens, Department of Engineering Management, University of Antwerp, Belgium. Corresponding author: Wannes Van den Bulck, wannes.vandenbulck@uantwerpen.be.}

\begin{abstract}
Large language models (LLMs) are increasingly used to translate technical explainable AI (XAI) artefacts such as SHAP into explanatory narratives (XAINs) for lay users. In high-risk domains, these explanations are often the main resource through which individuals can understand and contest adverse decisions. Yet LLMs are known to reproduce social bias, even from proxy variables alone, and fairness in XAINs remains unstudied. Because XAINs are bound to a ground-truth artefact, fairness concerns not only what a narrative says and how, but also whether it is equally faithful across social groups. We introduce fairness in XAINs as a new field of research at the intersection of fairness in XAI and fairness in natural language processing, and propose a framework that generates narratives under three conditions (protected attribute disclosed, withheld, or counterfactually swapped), allowing five comparisons that pinpoint the cause of disparities. We apply it to adversely classified instances from four datasets in credit scoring, education and employment, generating narratives with five LLMs from three providers, and evaluate faithfulness, feature coverage and sentiment. We find significant disparities across all datasets, metric families and models, arising in two ways. Faithfulness disparities result from feature values: LLMs make plausible but false statements, triggered by values that are unequally distributed across social groups. Feature coverage and sentiment disparities stem from the protected attribute itself: once it is disclosed, LLMs give less explanation to, or frame more negatively the narratives of, groups that have historically been discriminated against or are stereotypically associated with worse outcomes in the domain. A more faithful model removes the faithfulness disparities, and withholding protected attributes mostly removes the others, but neither suffices on its own. Fairness in XAINs should therefore be audited separately from the predictive model and its XAI artefact, using all three generation conditions.
\end{abstract}

\keywords{Explainable AI, XAI narratives, fairness, large language models, SHAP}

\maketitle

\section{Introduction}
\label{sec:intro}

As machine learning (ML) systems are increasingly used in high-risk domains, credit scoring, education assessment and employment screening stand out as cases where explainability is both legally required and socially important \citep{lo_piano_ethical_2020, fabeyo_explainable_2025, wang_explainable_2025, jang_practical_2022, kalasampath_literature_2025}. Because these models generally lack interpretability, the field of explainable artificial intelligence (XAI) aims to improve transparency by explaining why the AI system made a decision, allowing model developers and end-users to understand specific decisions and improve model trust \citep{molnar_interpretable_2020, saarela_recent_2024, adadi_peeking_2018}. 

This demand for transparency is not only socially important, but also increasingly of legal relevance: The EU AI Act classifies credit scoring, education and employment as high-risk applications under Annex III, meaning that under Article 86, any affected person subject to a decision taken on the basis of the output from such a high-risk system has the right to obtain a ``clear and meaningful explanation of the role of the AI system in the decision-making procedure and the main elements of the decision taken''\citep{euaiact2024}.
Among the XAI methods developed to meet the explainability demand, SHAP (SHapley Additive exPlanations) \citep{lundberg_unified_2017} has become the dominant approach \citep{saarela_recent_2024, kalasampath_literature_2025}. The SHAP algorithm quantifies how much each feature contributes to a model's prediction for a specific instance across all possible feature combinations and is widely regarded as the standard for post-hoc, instance-level explanations for tabular data. Despite its popularity, a problem remains: SHAP explanations are often too technical for applicants to understand or use \citep{rong_towards_2024, martens_tell_2025, kalasampath_literature_2025}, resulting in an \textit{interpretability gap} between what XAI methods produce and what decision subjects \textit{can} actually use. 

This has motivated the move toward eXplainable AI Narratives (XAIN), which use large
language models (LLMs) to translate technical XAI artefacts into natural language explanatory narratives
\citep{martens_tell_2025, silvestri_survey_2025}. The use of LLMs to translate XAI artefacts into narratives allows for structured, story-like explanations that establish cause-effect mechanisms between features \citep{cedro_importance_2026, martens_tell_2025}, improve comprehensibility, and are preferred by lay users over the raw artefacts they are derived from \citep{martens_tell_2025, silvestri_survey_2025}. An example of this translation step is given in Figure \ref{fig:narrativeintro}, of which the full narrative can be found in Appendix \ref{app:figures} in Figure \ref{fig:fullnarrative}.

\begin{figure}[!t]
\centering
\includegraphics[width=\linewidth]{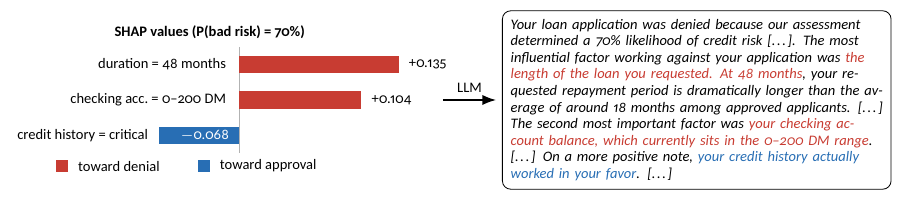}
\caption{From SHAP to XAIN: the top-3 SHAP features of a German Credit applicant (instance 18) and (part of) the narrative generated by \texttt{claude-sonnet-4-6}. Colours link each feature to its description in the narrative.}
\label{fig:narrativeintro}
\end{figure}

Yet the introduction of LLMs into the explanation pipeline introduces a new and unstudied risk. Coherent narrative generation requires access to the underlying feature values rather than SHAP scores alone \citep{ichmoukhamedov_how_2024, martens_tell_2025}, meaning at least parts of the applicant profile must be passed to
the LLM. In contexts qualified as high-risk, sensitive (protected) attributes are legally excluded from predictive model inputs and thus SHAP values will not be computed for them. But even if protected attributes (PAs) are not directly included in the generator prompt, proxies for these attributes are frequently present in the data and passed to the LLM as part of the applicant profile. 

As research consistently shows that state-of-the-art LLMs produce biased output even based on proxy variables alone
\citep{bai_explicitly_2025, gallegos_bias_2024, leidinger_how_2024}, there is no empirical or theoretical basis to expect LLM-generated narratives to be free from social biases. This motivates the research at hand. In high-risk settings where explanations are the main resource through which individuals can understand and improve adverse outcomes, disparities between the narratives of social groups risk allocational and representational harm \citep{gao_detecting_2026}, institutionalizing discrimination under the guise of transparency.

Despite these risks, fairness in XAINs remains unstudied. Fairness has been examined at every step of the pipeline (collecting data, making the prediction and explaining it), but not in the translation of the XAI artefact into a narrative (Fig.~\ref{fig:narrativeintro}): to the best of our knowledge, no study compares metrics computed on XAINs between social groups, nor do existing literature reviews on XAINs mention fairness concerns \citep{mill_real-world_2024, silvestri_survey_2025, bilal_llms_2025}. We therefore set aside fairness concerns in the earlier steps and focus exclusively on this translation step.

This paper addresses this gap with four contributions. First, we position fairness in XAINs as a distinct field of research at the intersection of fairness in XAI and fairness in natural language processing (NLP) (Section~\ref{sec:background}). Second, we propose a methodology for auditing fairness in XAINs that generates narratives under three conditions, in which protected attributes are factually disclosed, withheld, or replaced with counterfactual values, and evaluates them on faithfulness, feature coverage and sentiment. Comparing narratives across and within these conditions isolates total, indirect, disclosure, direct and value disparity (Section~\ref{sec:design}). Third, we present the first empirical evaluation of between-group differences in XAINs, generating narratives with five LLMs from three providers, ranging from small to frontier models, for adversely classified instances from four datasets covering the high-risk fields of credit scoring, employment and education. Fourth, we explain where each disparity comes from: the decomposition traces it to its source in the input (the applicant's feature values, the disclosure of the protected attributes, or their value), and inspecting the narratives behind it reveals the LLM behaviour that produces it (Section~\ref{sec:results}).\footnote{Code, prompts, narratives and extractions are available at \url{https://github.com/WannesVandenBulck/fairxain}.} Section~\ref{sec:discussion} discusses the implications of these findings, and Section~\ref{sec:conclusion} concludes.

\section{Background}
\label{sec:background}

\subsection{Explainable AI and the Interpretability Gap} 

The XAI literature is dominated by post-hoc methods in tabular data that explain individual model predictions \citep{silvestri_survey_2025, kalasampath_literature_2025}. The field grew from the need to explain black-box ML models, after their increasing effect on human lives. Among these, feature attribution artefacts like SHAP \citep{lundberg_unified_2017} and LIME (Local Interpretable Model-agnostic Explanations) \citep{ribeiro_model-agnostic_2016} are the most popular, of which SHAP has become the most widely adopted approach \citep{saarela_recent_2024, kalasampath_literature_2025}. However, like most post-hoc explanation methods, research shows that such explanations are too technical for lay people to interpret \citep{rong_towards_2024}, a failure \citet{miller_explainable_2017} describe as ``inmates running the asylum,'' in which XAI methods are designed by and for data scientists instead of for the individuals whose lives the underlying decisions affect \citep{bhatt_explainable_2020}. This results in an interpretability gap that sits in direct contrast with the purpose of the field: if XAI was introduced to make AI decisions understandable to those affected by them, then methods that remain uninterpretable result in perverse outcomes.

In the high-risk domains studied in this paper, all those affected are by definition lay users: loan and job applicants, and students. They cannot be expected to interpret what a SHAP value such as $+0.135$ means and how the features compare (Fig.~\ref{fig:narrativeintro}). This results in the inability of adversely classified individuals to understand and contest the decision.  

\subsection{Explainable AI Narratives}
\label{explainable ai narratives}

XAINs address the interpretability gap by translating technical XAI artefacts into natural language explanations through LLMs \citep{martens_tell_2025}. An XAIN presents a structured, story-like explanation, including cause-effect relationships between input features and the model's decision \citep{cedro_importance_2026}. This is an advancement from earlier approaches, which were restricted to restating attribution scores or feature-value pairs in prose form, without organizing them into a coherent, context-sensitive narrative \citep{cambria_survey_2023}, distinguishing XAI narratives from XAI descriptions \citep{cedro_importance_2026}. \citet{martens_tell_2025} introduced SHAPstories as a narrative counterpart to SHAP explanations, showing substantial improvements in comprehensibility and preference by both lay users and domain experts. 

XAINs are generated through a unified \citep{martens_tell_2025, ichmoukhamedov_how_2024} pipeline (Figure \ref{fig:XAINpipeline}), which goes as follows. The process begins with a trained ML model producing a prediction together with the corresponding decision. SHAP values are then computed on top of this prediction and ranked by absolute importance to obtain the top contributing features. This is combined with several additional components (laid out in Section \ref{llm-providers}) into a structured prompt (Fig. \ref{fig:generator_prompt}), which is passed to an LLM to generate the XAIN. 

\subsection{Fairness in XAI and NLP}

Fairness has been studied separately within the fields of XAI and NLP. Explanation fairness in the context of XAI studies whether the explanations produced by post-hoc XAI artefacts such as SHAP are consistent or of comparable quality across social groups \citep{zhao_fairness_2023, dai_fairness_2022, balagopalan_road_2022}, auditing the fidelity, stability, consistency and sparsity. The field of fairness in XAI seeks to provide explanations of equal quality to different social groups, but until now has left open whether these artefacts are translated fairly into natural-language narratives. 

Meanwhile, the NLP literature has documented how LLMs encode and reproduce social bias. Foundational works have shown that LLMs absorb and amplify stereotypical associations present in their training corpora \citep{bolukbasi_man_2016, caliskan_semantics_2017}, and more recent works confirm that LLMs continue to exhibit biases across a wide range of generation tasks \citep{gallegos_bias_2024}. This bias persists even when protected attributes are not present in the input, as LLMs rely on proxy variables correlated with them \citep{bai_explicitly_2025, leidinger_how_2024}.

The consequences of such disparities are of allocational harm, referring to the unequal distribution of resources or opportunities (e.g., access to information on how to escape the adverse class), and representational harm, referring to the reinforcement of stereotypes or the framing of certain groups in persistently more negative terms \citep{gao_detecting_2026}. Both are also relevant to XAINs, since a narrative can misrepresent an individual's situation and affect their ability to act on or contest an adverse decision.

Despite this, neither the XAI fairness literature nor the NLP fairness literature directly addresses the setting most relevant to XAINs: fairness in instance-specific, natural-language explanations generated for adversely predicted individuals. Current works in fairness in XAI evaluate the technical artefact but not the XAIN that translates it, while works in fairness in NLP typically study bias in open-ended generation, instead of the constrained, artefact-grounded generation that characterizes XAINs. Because an XAIN is grounded in an XAI artefact, that artefact constitutes a ground truth to which the narrative must remain faithful. Fairness in XAINs therefore cannot be assessed on the content of a narrative alone, such as which features it mentions (feature coverage) or which sentiment it contains (sentiment), but must also account for whether narratives accurately translate this ground truth (faithfulness). This introduces a fairness dimension without an analogue in either parent field. Consequently, no framework currently exists to assess disparities in XAINs across social groups \citep{mill_real-world_2024, silvestri_survey_2025}, motivating this work.

\section{Study Design}
\label{sec:design}

In this section, the study design will be presented. First, an overview is given of the datasets and models used to make the predictions and compute the SHAP features for the instances for which the XAINs provide explanations. Second, the choice of generator LLMs and the prompt design are justified. Third, the different generation conditions are laid out. Lastly, the metrics on which social groups will be compared are presented.

\subsection{Datasets and Models}
\begin{figure}[!t]
  \centering  \includegraphics[width=0.65\linewidth]{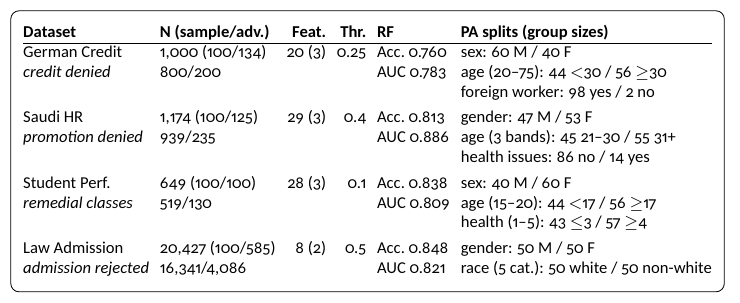}
 \caption{Overview of the datasets. \textit{N (sample/adv.)}: total instances, with in brackets the number of sampled instances and of adversely predicted test instances, and the train/test split below. \textit{Feat.}: number of features, of which protected attributes in brackets. \textit{Thr.}: predicted-probability threshold for an adverse prediction. \textit{RF}: test accuracy and AUC of the Random Forest. \textit{PA splits}: value range among the sampled instances in brackets, followed by the two groups compared and their sizes. Saudi age bands: 21--30, 31--40, 41+; student health: 1 (very bad), 2 (bad), 3 (fair), 4 (good), 5 (very good); law race: white, Hispanic, Asian, Black, other.}
  \label{fig:datasets}
  \Description{Overview of used datasets and their characteristics.}
\end{figure}

We evaluate our framework on four publicly available datasets drawn from high-risk domains as defined by the EU AI Act \citep{euaiact2024}: (i) German Credit \citep{statlog_german_credit_data_144}, recording loan applications from 1970s Germany, for credit scoring; (ii) Saudi Employee Turnover (Saudi HR) \citep{alqahtani2025saudi}, collected through a survey of employees of Saudi Arabian private-sector companies, for employment; (iii) Portuguese Student Performance (Student) \citep{student_performance_320}, recording secondary-school students of the Portuguese language course at two Portuguese schools; and (iv) Law School Bar Passage (Law) \citep{ofer_lawschool_kaggle}, drawn from the LSAC National Longitudinal Bar Passage Study \citep{wightman1998lsac}, both for education. For each dataset, a Random Forest (RF) classifier with balanced class weights was trained on an 80/20 train/test split, excluding the protected attributes, to predict (i) bad credit risk, (ii) employee attrition, (iii) failing the final year of school and (iv) failing the bar exam. Each prediction is linked to an adverse decision: an applicant with bad credit risk is denied credit, and, hypothetically, an employee predicted to leave is denied a promotion, a student predicted to fail is assigned mandatory remedial classes, and an applicant predicted to fail the bar exam is rejected for law school admission. Decision thresholds were set to result in enough adversely predicted test instances for statistical analysis. From these, 100 instances per dataset were sampled, spread as evenly as possible over the combinations of protected-attribute groups so that small groups are fully represented. For each sampled instance, SHAP feature attributions were computed with TreeSHAP \citep{lundberg_unified_2017}: a SHAP value quantifies a feature's effect on the prediction by averaging its contribution across all possible feature combinations. Figure~\ref{fig:datasets} summarises each dataset, its model and the compared groups.

\subsection{Generator LLMs and Prompt Design}
\label{llm-providers}

We generate XAINs with five LLMs from three providers: \texttt{gpt-4.1} and \texttt{gpt-4o-mini} (OpenAI), \texttt{claude-sonnet-4-6} (Anthropic), and \texttt{mistral-large-2512} and \texttt{ministral-14b} (Mistral AI). They differ in size, weight availability, and region of origin. An overview is given in Table~\ref{tab:llms_overview} (Appendix~\ref{app:tables}). All models are called at temperature $T = 0$ to minimise variation between repeated calls and support reproducibility. 

Each LLM receives an instance-specific prompt, of which the template is given in Fig. \ref{fig:generator_prompt} in Appendix \ref{app:figures}. The prompt consists of five parts, preceded by the explanation goal: (i) the dataset context, describing the data and the target variable and stating that the protected attributes were not used by the model; (ii) a description of what SHAP values mean; (iii) the applicant profile, listing each feature value with its description and the average or distribution among the accepted group (the test instances predicted to receive the favourable outcome), followed by the predicted probability and the decision threshold; (iv) the top-3 features ranked by absolute SHAP value, with their SHAP values; and (v) the narrative task, specifying what the narrative must explain, together with constraints on content and style. This prompt structure follows and extends the pipeline proposed by \citet{martens_tell_2025} and similarly used in other works evaluating narratives \citep{ichmoukhamedov_how_2024, he_agentic_2026}. One notable difference is the inclusion of feature averages and distributions of the accepted group next to the feature value of the adversely predicted instance. In doing this, we follow \citet{he_agentic_2026} and \citet{ichmoukhamedov_how_2024}, but only include averages and distributions for the accepted group (instead of the dataset as a whole), as this has been deemed of greater relevance. 

Protected attributes could be included as part of the instance profile passed to the LLM. This choice can be motivated in several ways. First, coherent narrative generation requires access to the underlying feature values: an LLM cannot construct a clear, cause-effect explanatory account of an individual's situation without knowledge of their actual feature values, to which the inclusion of protected attributes may add. Second, including this information can support personalisation objectives, allowing the narrative to connect model reasoning to the individual's specific situation rather than producing a generic explanation \citep{musicco_who_2026}. Third, in certain application domains (e.g. medical contexts), the disclosure of attributes such as age or sex may be necessary for the explanation to be meaningful or actionable. Finally, preliminary considerations suggest that certain metrics may perform differently depending on whether such attributes are visible to the model \citep{bulck26a}. Given these considerations, we treat the disclosure of protected attributes as an experimental variable, as will be detailed in Section~\ref{generation_conditions}.

\subsection{Generation Conditions}
\label{generation_conditions}

\begin{figure}[!t]
  \centering
  \includegraphics[width=0.65\linewidth]{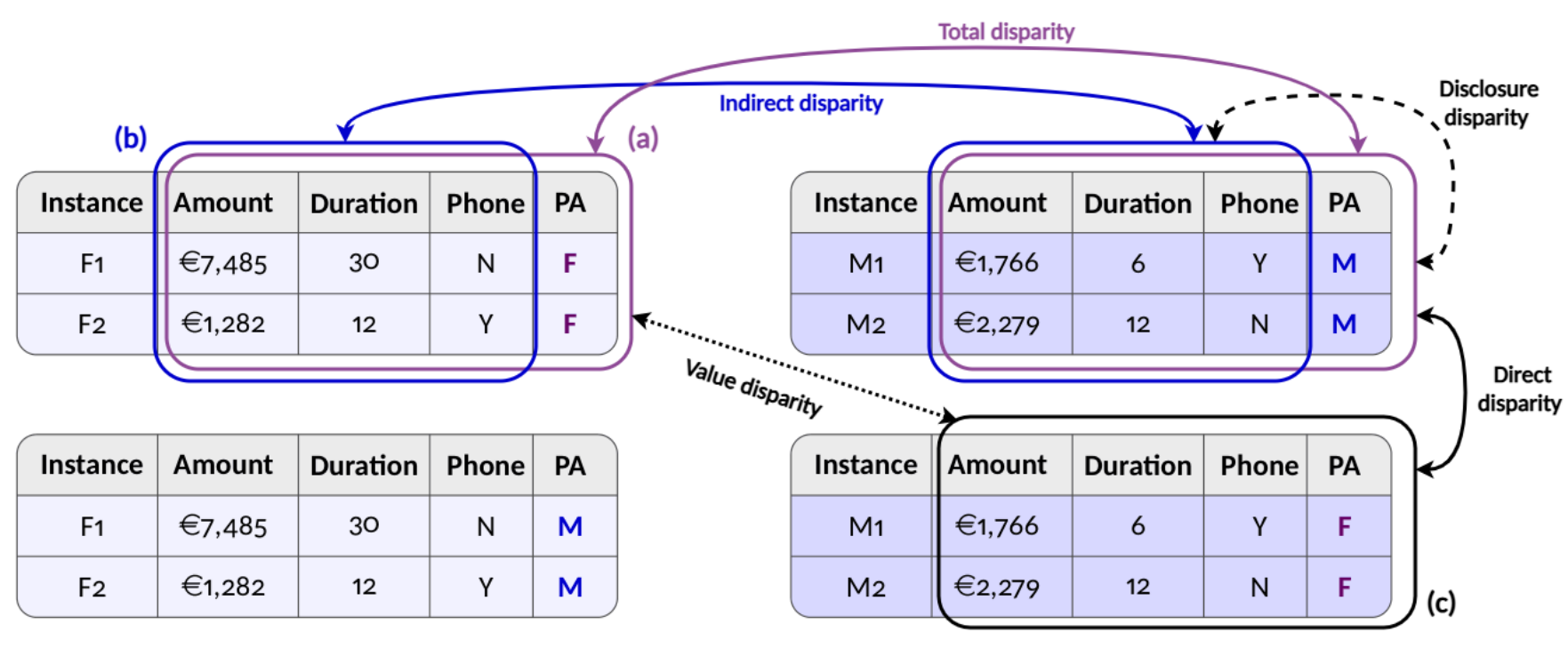}
  \caption{Illustration of the three narrative generation conditions and the five resulting disparity comparisons, based on real (top) and counterfactual PA-swapped (bottom) credit-dataset instances for female (F) and male (M) applicants.}
  \label{fig:methodology}
  \Description{Illustration of the three narrative generation conditions and the five resulting disparity comparisons, based on real (top) and counterfactual PA-swapped (bottom) credit-dataset instances for female (F) and male (M) applicants.}
\end{figure}

Building on the XAIN generation pipeline described in Section~\ref{explainable ai narratives}, we vary the prompt's protected attribute content to isolate distinct sources of disparity. We generate narratives for the adversely classified instances under three conditions, defined by the combination of two prompt-level factors: protected attribute type (factual vs. counterfactual) and protected attribute visibility (disclosed vs. withheld). A factual protected attribute matches the instance's actual group membership, whereas for a counterfactual protected attribute the entire feature vector is held fixed while the value of the protected attribute is swapped. This gives three generation conditions, illustrated by Figure~\ref{fig:methodology}:

\begin{itemize}
    \item[(a)] \textbf{Factual PA disclosed}: the instance's true protected attribute is included in the prompt.
    \item[(b)] \textbf{Factual PA withheld}: the instance's true protected attribute is removed from the prompt.
    \item[(c)] \textbf{Counterfactual PA disclosed}: the protected attribute is disclosed but swapped with another value, with the remaining feature vector held fixed.
\end{itemize}

Each protected attribute is swapped one at a time to a fixed value of the other group: binary attributes take the other value, ordinal attributes the median of the other group (age: 38 or 25 in German Credit, 31--40 or 21--30 in Saudi HR, 17 or 15 in Student Performance; student health: ``very good'' or ``bad''), and race its most common value (black for white applicants, white for non-white applicants). We define a disparity as a statistically significant difference on any of the metrics introduced in Section~\ref{metrics},  between social groups or between generation conditions. Comparing the conditions gives five types of disparity (Figure~\ref{fig:methodology}):

\begin{itemize}
    \renewcommand{\labelitemi}{--}
    \item \textbf{D1: Total disparity}: comparing social groups under condition (a) shows whether a disparity exists at all, combining feature-driven and label-driven effects.
    \item \textbf{D2: Indirect disparity}: comparing social groups under condition (b) shows whether a disparity persists when the PAs are hidden from the LLM, attributing the disparity to feature values.
    \item \textbf{D3: Disclosure disparity}: contrasting (a) with (b) within a social group shows whether disclosing the group's protected attribute changes the narrative.
    \item \textbf{D4: Direct disparity}: contrasting (c) with (a) for the same instance, with only one protected attribute swapped to another value, isolates the effect of the PA value itself.
    \item \textbf{D5: Value disparity}: contrasting (c) with (a) for narratives that display the same PA values, but whose feature values come from different social groups, shows whether the feature values cause a disparity even when the LLM sees the same label. Unlike indirect disparity, this captures feature effects that only arise when a protected attribute is visible. Narratives that display the same values for all protected attributes are compared.
\end{itemize}

Together, these comparisons detect disparities and isolate their cause: the feature values (D2, D5), the disclosure of the protected attribute (D3), or its value (D4).

\subsection{Metrics}
\label{metrics}

In this section, the metrics on which the comparisons will be made are presented. We compute faithfulness to the underlying XAI artefact (which is the most frequently evaluated criterion \citep{silvestri_survey_2025}), feature coverage, and sentiment. A full overview of all (individual) metrics is given in Table \ref{tab:metrics}. 

\textit{(i) Faithfulness}. The generator LLM is presented with the top $X$ features based on the absolute SHAP values, with $X = 3$ in this study. Following the taxonomy of \citet{silvestri_survey_2025}, the narrative should be accurate and complete in representing the underlying XAI artefact. Accuracy in this context is split into three parts \citep{ichmoukhamedov_how_2024}. Value Accuracy (VA) measures whether the stated feature values are correct, Sign Accuracy (SA) whether the direction of each feature's influence on the prediction is correct, and Rank Accuracy (RA) whether the most important features are presented in the correct order of importance.

For SHAP faithfulness, these metrics measure the agreement with the ground-truth values $v_j^*, s_j^*, r_j^*$, representing the actual feature value, sign and rank of the extracted feature $f_j$. For each top-$X$ feature, a binary indicator records whether the extracted quantity is correct (Table~\ref{tab:metrics}). Per narrative, these indicators are aggregated into VA, SA and RA, each computed as an accuracy (XA) of the same general form \citep{ichmoukhamedov_how_2024} (Equation~\ref{eq:xa}), where $x_j$ denotes the extracted quantity (value, sign, or rank) for feature $f_j$, and $x_j^*$ its ground-truth counterpart. The Kronecker delta $\delta_{x_j, x_j^*}$ equals 1 if the extracted quantity matches the ground truth and 0 otherwise, so the numerator sums the number of correct matches. The denominator normalizes this count by the number of scored features $n$ minus those for which no quantity could be identified in the narrative (denoted $x_j = \phi$). In other words, XA is the accuracy of the extracted quantity, computed only over features where the narrative provided an extractable value:
\begin{equation}
XA = \sum_{x_j \neq \phi} \frac{\delta_{x_j, x_j^*}}{n - \sum_{x_j = \phi} 1}.
\label{eq:xa}
\end{equation}

Value accuracy is scored in the same way for any secondary features (features outside the top-$X$ SHAP features) and protected attributes the narrative mentions, and the per-narrative VA aggregates all of these. Finally, we also record whether the predicted probability is stated correctly.

\textit{(ii) Feature Coverage}. In addition to the correctness of the included features, it must be ensured that the narrative includes all that it needs to include. Completeness here is defined as the presence of each of the top-$X$ SHAP features in the narrative. Likewise, the narrative may include secondary features if the LLM deems these relevant. All features mentioned in the narrative are therefore extracted, and for each feature a binary indicator records whether it is mentioned. This allows for comparisons of completeness (for the top-$X$ SHAP features) and general feature coverage (for secondary features). The protected attributes are treated separately: since they were excluded from model training, they never carry a SHAP value, and any mention of them is introduced by the LLM itself. Per narrative, we summarize coverage in two proportions: the \textit{PA-mention proportion}, the share of the dataset's protected attributes that are mentioned, and the \textit{secondary-feature-mention proportion}, the share of features outside the top $X$ that are mentioned.

\textit{(iii) Sentiment}. Beyond what a narrative states, fairness also concerns how it is phrased: two narratives can be equally faithful and complete, yet one may sound encouraging and the other discouraging. We measure the tone of each narrative with the Bing Liu opinion lexicon \citep{lexicon}, a list of 2,006 positive and 4,783 negative English words widely used in sentiment research. Each narrative is lowercased and split into words, and its net sentiment is computed as
\begin{equation}
S = \frac{n_{pos} - n_{neg}}{n_{pos} + n_{neg}},
\label{eq:sentiment}
\end{equation}
where $n_{pos}$ and $n_{neg}$ are the number of positive and negative words, so that $S$ ranges from $-1$ (only negative words) to $1$ (only positive words), and is set to 0 when the narrative contains none. Sentiment is computed on the full narrative. 

Each metric shows a different way in which narratives could differ between social groups. Faithfulness could differ if LLMs unequally flip the sign of SHAP values they deem ``counterintuitive'' \citep{ichmoukhamedov_how_2024}, or alter feature values to fit their expectations across groups. Feature coverage could differ because LLMs associate social groups with certain characteristics \citep{bai_explicitly_2025}, so a protected attribute or a group's typical profile may change what the LLM deems relevant to mention. Sentiment could differ if the LLM frames groups according to stereotypes it absorbed from its training data \citep{gallegos_bias_2024}. Whether, and why, each of these occurs is shown in Section~\ref{sec:results}.

\subsection{Evaluation Pipeline}
\label{evaluation}

\begin{figure}[!h]
  \centering
  \includegraphics[width=0.6\linewidth]{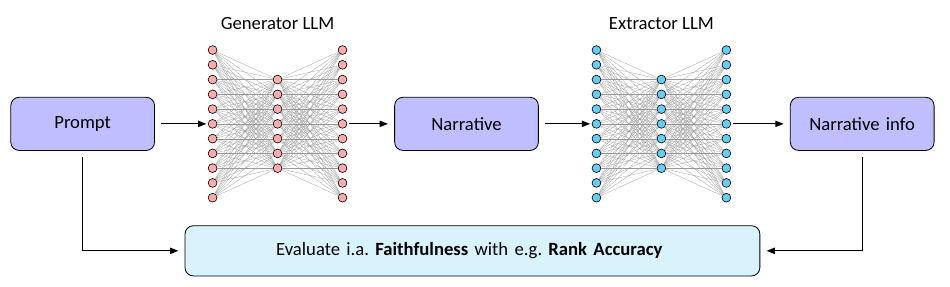}
  \caption{Theoretical representation of the evaluation pipeline as performed in this paper}
  \label{fig:evalpipeline}
  \Description{Theoretical representation of the evaluation pipeline as performed in this paper}
\end{figure}

The faithfulness and feature coverage metrics are computed using the pipeline depicted in Figure ~\ref{fig:evalpipeline}. 
We adopt an extraction-based evaluation pipeline, in line with earlier works \citep{ichmoukhamedov_how_2024, he_agentic_2026}. This pipeline extracts information from each narrative into structured information in JSON format, which can then be compared against the ground truth from the generation prompt. 

Because of budget constraints, all narratives are processed by a single extractor LLM, \texttt{grok-4.20-0309-non-reasoning}, which receives the narrative without access to the ground truth its extraction is compared with. We chose an extractor from a provider that generated none of the narratives, as LLMs tend to favour text produced by their own model family. The extraction prompt format is shown in Figure~\ref{fig:extractionprompt}. To validate the extractor, we manually checked the extractions of 100 randomly sampled narratives (five per generator LLM and dataset). The extraction was fully correct for 94 narratives, the six errors (three values, one sign, one top-3 selection and one missed probability) were spread across generator LLMs, showed no pattern by social group, and are in line with earlier research \citep{ichmoukhamedov_how_2024}. For every statistically significant disparity (Section~\ref{sec:results}), we verified that no extraction errors were made. The remaining risk is that the extractor misses errors, which could hide disparities, although \citet{ichmoukhamedov_how_2024} found such misses to never occur. 

The extraction is compared against a ground truth file, derived directly from the generation prompt, which provides an objective reference against which the extractions can be compared. 

\section{Results}
\label{sec:results}

Results are presented per metric family: faithfulness, feature coverage and sentiment. For each, we first describe the general picture and then the differences between social groups. To locate where a difference comes from, we use the five disparity comparisons of Section~\ref{generation_conditions} (D1--D5): a total disparity (D1) shows that groups differ, and the other four show whether this is due to the feature values (D2, D5), the disclosure of the protected attribute (D3) or its value (D4). Inspecting the narratives behind each disparity then reveals the LLM behaviour that produces it. Each metric was tested separately per dataset and generator LLM, and further per protected attribute (D1, D2), per social group (D3) or per counterfactual swap (D4, D5), using two-sided tests that match each design: independent-group tests for D1 and D2 (Appendix~\ref{app:independent}), paired tests for D3 and D4 (Appendix~\ref{app:paired}) and stratified tests for D5 (Appendix~\ref{app:stratified}). Raw $p$-values were corrected with the Benjamini--Hochberg procedure ($\alpha = 0.05$) within batches that keep binary and continuous metrics apart (Appendix~\ref{app:stats-general}, Table~\ref{tab:fdr-batches}). Only results that remain significant after this correction are reported as findings.

For brevity, this section focuses on the most notable findings, but all significant findings are to be found in Tables \ref{tab:faith_findings}, \ref{tab:cov_findings} and \ref{tab:sent_findings}. Across all three metrics, two ways in which disparities occur emerge. Faithfulness disparities arise from the feature values: the LLM makes errors that are blind to the social group but triggered by values that are unequally distributed across groups. Disparities in feature coverage and sentiment arise mainly from the protected attribute itself: whether it is disclosed, and which value it takes, changes the narrative's content. 

\subsection{Faithfulness}

\subsubsection{General picture}
Faithfulness follows model cost (Table~\ref{tab:faith_general} and Table~\ref{tab:llms_overview}; tests in Appendix~\ref{app:model-level}). The two most expensive models are the most faithful: \texttt{claude-sonnet-4-6} leads on sign and rank accuracy, \texttt{gpt-4.1} on value accuracy. The two small models trail far behind, especially on sign accuracy (71--76\%). The fewest mistakes are made for value accuracy: every model states at least 90\% of feature values correctly on every dataset. The predicted probability of the adverse outcome is stated correctly in all narratives, except for \texttt{ministral-14b} on Saudi HR (90.5\%).

\begin{table}[!h]
\footnotesize
\setlength{\tabcolsep}{3pt}
\caption{Faithfulness per generator LLM and dataset under the $PA_{\text{disclosed}}$ condition (\%): value (VA), sign (SA) and rank (RA) accuracy.}
\label{tab:faith_general}
\begin{tabular}{@{}l ccc ccc ccc ccc ccc@{}}
\toprule
 & \multicolumn{3}{c}{German Credit} & \multicolumn{3}{c}{Saudi HR} & \multicolumn{3}{c}{Student} & \multicolumn{3}{c}{Law} & \multicolumn{3}{c}{All} \\
\cmidrule(lr){2-4}\cmidrule(lr){5-7}\cmidrule(lr){8-10}\cmidrule(lr){11-13}\cmidrule(l){14-16}
Model & VA & SA & RA & VA & SA & RA & VA & SA & RA & VA & SA & RA & VA & SA & RA \\
\midrule
\texttt{claude-sonnet-4-6}  & 98.9 & 98.0 & 84.0 & 99.9 & 99.3 & 96.3 & 96.9 & 97.7 & 97.7 & 99.8 & 99.3 & 91.7 & 98.9 & \textbf{98.6} & \textbf{92.4} \\
\texttt{gpt-4.1}            & 100  & 92.7 & 93.0 & 100  & 98.0 & 81.0 & 99.5 & 91.7 & 93.7 & 99.7 & 95.3 & 94.7 & \textbf{99.8} & 94.4 & 90.6 \\
\texttt{mistral-large-2512} & 90.0 & 81.0 & 72.3 & 98.0 & 92.3 & 85.0 & 95.3 & 88.0 & 83.3 & 98.2 & 97.3 & 81.3 & 95.0 & 89.7 & 80.5 \\
\texttt{gpt-4o-mini}        & 96.1 & 60.5 & 74.0 & 94.6 & 64.0 & 87.3 & 93.0 & 71.0 & 57.3 & 99.5 & 89.2 & 71.3 & 95.5 & 71.2 & 72.5 \\
\texttt{ministral-14b}      & 91.7 & 67.8 & 70.3 & 93.0 & 71.7 & 79.3 & 93.5 & 72.1 & 63.0 & 93.7 & 92.8 & 78.3 & 92.9 & 76.2 & 72.8 \\
\bottomrule
\end{tabular}
\end{table}

German Credit has the lowest average sign accuracy, mainly because three SHAP-favourable values sound unfavourable, but aren't: no checking account, no or unknown savings, and a ``critical'' credit history (43\% of its sign errors). \texttt{gpt-4o-mini} never describes them correctly (\texttt{claude-sonnet-4-6}, \texttt{gpt-4.1}: 80--94\%). This illustrates the most frequent kind of error, which we call a \textit{plausible falsehood}: a statement that sounds reasonable but contradicts the SHAP artefact. Most commonly, a favourable feature (one pushing toward the favourable outcome) is presented as a reason for the adverse decision, which accounts for 86\% of all sign errors. Smaller models more often include these \textit{plausible falsehoods}: 51--59\% of favourable features are described this way, in comparison with 3\% for \texttt{claude-sonnet-4-6}. Mistakes against the rank accuracy are mostly reorderings of the correct top-3 SHAP features (82\%). Including a secondary feature into the top three also happens almost exclusively in the smallest two models (91\% of cases).

\subsubsection{Differences between social groups}
None of the 19 faithfulness disparities (as shown in Table \ref{tab:faith_findings}) occur in \texttt{claude-sonnet-4-6} or \texttt{gpt-4.1}, mainly because they correctly handle the cases that cause disparities in smaller models, so many metrics are correct for every narrative and disparities between social groups do not show up.

In the smaller models, all disparities follow the same plausible falsehood principle. Errors are triggered by particular feature values, but because these values are unequally distributed across groups, so are by extension the errors. By inspecting the narratives that include these errors, we identify two mechanisms: (i) the LLM's prior expectations override the artefact, or (ii) the LLM bends the facts to justify the adverse decision.

(i) The clearest case is the work experience feature in Saudi HR: 1--5 years of experience is favourable in every instance, yet \texttt{gpt-4o-mini} reverses this in 37 of 39 cases, reasoning that greater experience \textit{``correlates with stability and commitment''}. As 53\% of women hold 1--5 years of experience against 26\% of men, and 69\% of employees aged 21--30 against 16\% of those aged 31+, the sign of the most important feature is correct for only 37\% of women against 70\% of men (D1, $p_{FDR}=0.019$), and for 34\% of young against 67\% of older employees (D1, $p_{FDR}=0.024$). The decomposition confirms that the feature values, not the protected attribute, drive this: relabelling young employees as older leaves their accuracy unchanged (34\% to 32\%; D4, $p=1.0$), while their profiles still fall far behind those of actually older employees (31\% against 68\%; D5, $p_{FDR}=0.026$). Similarly, \texttt{gpt-4o-mini} ranks checking-account status first in German Credit regardless of its SHAP rank, seemingly following domain expectation. As women less often have this as their actual top feature (38\% against 57\%), their top feature is identified correctly less often (65\% against 90\%; D1, $p_{FDR}=0.040$).

(ii) When nothing in the prompt seems to support the adverse outcome, the LLM invents a reason. In 33\% of Student Performance instances, all three top SHAP features push toward passing, and both \texttt{ministral-14b} and \texttt{mistral-large-2512} then wrongly turn favourable features into risks, e.g. \textit{``your record shows you've faced challenges in the past''} for students without a single failed class. Students in poor health are more often in this situation (44\% against 25\%), producing an indirect disparity (\texttt{mistral-large-2512}: 92\% against 83\%; D2, $p_{FDR}=0.044$; \texttt{ministral-14b}: 75\% against 56\%; D2, $p_{FDR}=0.045$). In Saudi HR, \texttt{gpt-4o-mini} goes further and replaces an employee's value with the most common value among accepted employees, then adapts the explanation to fit: employees with outside job offers are told they \textit{``have had limited external offers''}, which \textit{``might suggest a lack of confidence in your current role''}. This behaviour becomes more frequent when the protected attributes are withheld (20 against 6 of 38 cases). Women and employees aged 21--30 are affected most: their value accuracy is lower when the protected attributes are withheld than when they are disclosed (women: 84.2\% against 95.6\%; aged 21--30: 81.5\% against 94.8\%; D3, $p_{FDR}=0.010$ and $0.025$).

Taken together, faithfulness disparities mostly (but not exclusively, see Table \ref{tab:faith_findings}) disadvantage women and young employees, and are driven by feature values, not by the protected attributes. We show that withholding the protected attributes therefore does not remove them (D2), and changing their value hardly matters (D4). Whether the protected attributes are disclosed can even make things worse, in either direction: withholding them made \texttt{gpt-4o-mini} fabricate more values for women and young employees, while disclosing them made \texttt{ministral-14b} misstate the experience sign more often for young employees.

\subsection{Feature Coverage}

\subsubsection{General picture}
All models mention nearly all top-3 SHAP features, and therefore produce complete XAINs (Table~\ref{tab:cov_general}). Only the two small models occasionally forget one, most often in German Credit (91.0--94.7\%). Protected attributes are rarely mentioned (1.1--6.8\%), and outside of Student Performance, 45 of 46 mentions are non-discrimination disclaimers saying that the protected attributes were not used by the predictive model. In addition to these disclaimers, protected attributes are mentioned only in Student Performance for the health feature: narratives from \texttt{gpt-4o-mini}, \texttt{ministral-14b} and \texttt{mistral-large-2512} use the student's health in the explanation, either presenting poor health as a cause or good health as a strength, although the prompt states that health was not used by the predictive model. 

\begin{table}[!h]
\footnotesize
\setlength{\tabcolsep}{3pt}
\caption{Feature coverage per generator LLM and dataset under the $PA_{\text{disclosed}}$ condition (\%), averaged over narratives. C: share of the top-3 SHAP features mentioned in a narrative. PA: share of the dataset's protected attributes mentioned in a narrative. Sec.: share of the dataset's secondary features mentioned in a narrative.}
\label{tab:cov_general}
\begin{tabular}{@{}l ccc ccc ccc ccc ccc@{}}
\toprule
 & \multicolumn{3}{c}{German Credit} & \multicolumn{3}{c}{Saudi HR} & \multicolumn{3}{c}{Student} & \multicolumn{3}{c}{Law} & \multicolumn{3}{c}{All} \\
\cmidrule(lr){2-4}\cmidrule(lr){5-7}\cmidrule(lr){8-10}\cmidrule(lr){11-13}\cmidrule(l){14-16}
Model & C & PA & Sec. & C & PA & Sec. & C & PA & Sec. & C & PA & Sec. & C & PA & Sec. \\
\midrule
\texttt{claude-sonnet-4-6}  & 99.7 & 0.0 & 8.4  & 100  & 1.0 & 5.3  & 100  & 1.0 & 4.4  & 100  & 25.0 & 20.3 & 99.9 & 6.8 & 9.6 \\
\texttt{gpt-4.1}            & 100  & 0.0 & 0.4  & 99.7 & 6.0 & 0.2  & 100  & 0.0 & 1.3  & 100  & 3.0  & 8.0  & 99.9 & 2.2 & 2.5 \\
\texttt{mistral-large-2512} & 98.3 & 0.0 & 8.3  & 100  & 2.0 & 11.8 & 100  & 1.7 & 12.6 & 100  & 5.0  & 24.0 & 99.6 & 2.2 & 14.2 \\
\texttt{gpt-4o-mini}        & 94.7 & 0.0 & 1.9  & 99.0 & 0.0 & 3.2  & 95.7 & 4.3 & 3.9  & 98.7 & 0.0  & 10.3 & 97.0 & 1.1 & 4.8 \\
\texttt{ministral-14b}      & 91.0 & 1.3 & 10.2 & 99.7 & 0.0 & 9.8  & 93.7 & 3.0 & 7.5  & 93.0 & 1.0  & 23.3 & 94.3 & 1.3 & 12.7 \\
\bottomrule
\end{tabular}
\end{table}

\subsubsection{Differences between social groups}
Unlike faithfulness, significant differences in feature coverage also occurred in \texttt{claude-sonnet-4-6},  accounting for 17 of the 34 findings, all summarised in Table \ref{tab:cov_findings}. Here, disparities are mainly driven by the protected attribute itself, through its value and its disclosure. We identify two mechanisms: (i) the protected attribute prompts the model to talk about the attribute itself, or (ii) favourable feature values are brought in as strengths.

(i) When a protected attribute is disclosed, some models comment on it, and how often depends on its value. \texttt{claude-sonnet-4-6} adds non-discrimination disclaimers in Law (\textit{``It is important to note that your gender and race played no role in this decision''}) far more often for non-white than for white applicants (38\% against 12\%; D1, $p_{FDR}=0.016$), and most often for Hispanic applicants (60\%). This is not explained by their higher predicted failure risk, but by the race value itself: relabelling non-white applicants as white reduces it from 38\% to 10\% (D4, $p_{FDR}=0.012$). In Student Performance, \texttt{gpt-4o-mini} mentions health almost only when it is poor (27.9\% against 1.8\%; D1, $p_{FDR}=0.003$). Nearly half of these mentions claim that health affects the outcome, e.g. \textit{``your current health status is rated as very bad, which can significantly affect your ability to focus and perform well in school''}, although the prompt states that health was not used by the model. For these students, including the bad health label raises health mentions from 2\% to 44\% (D4, $p_{FDR}<0.001$). In both cases, withholding the protected attributes removes the mentions almost entirely (D3). The PA mentions bring another disadvantage, as they take up space. Within the 250-word limit, \texttt{claude-sonnet-4-6}'s Law narratives that contain a non-discrimination disclaimer mention fewer secondary features (10.7\% against 23.6\% without one). As a result, non-white applicants, who receive them most, get less explanation when race is disclosed: their narratives mention 14.7\% of the secondary features on average, against 28.0\% when race is withheld (D3, $p_{FDR}=0.025$).

(ii) Some disparities follow from the applicant's profile: models tend to mention a secondary feature when its value is favourable. In Law, an undergraduate GPA above the accepted average is mentioned in 45\% of narratives against 25\% when below it, and women more often have such a GPA. Men's profiles labelled as female receive far fewer GPA mentions than actual women (\texttt{claude-sonnet-4-6}: 13\% against 54\%; D5,  $p_{FDR}=0.022$; \texttt{gpt-4o-mini}: 3\% against 43\%; D5,  $p_{FDR}=0.013$), confirming that the label does not cause this. Similarly, with age withheld, \texttt{gpt-4o-mini} mentions more secondary features for older German Credit applicants (3.1\% against 1.0\%; D2, $p_{FDR}=0.042$), mostly employment duration and credit history, which tend to be longer for older applicants.

Coverage disparities thus mostly (but not exclusively, see Table~\ref{tab:cov_findings}) disadvantage non-white applicants, who receive more non-discrimination disclaimers at the cost of secondary feature mentions, and students in poor health, who are wrongly told that their health negatively affects the outcome. Unlike faithfulness disparities, they are driven mainly by the protected attributes. Withholding the protected attributes therefore removes these disparities, leaving only those driven by feature values, when these values seem favourable.

\subsection{Sentiment}

\subsubsection{General picture}
All narratives have a positive net sentiment on average (Table~\ref{tab:sent_general}), despite explaining an adverse decision. \texttt{claude-sonnet-4-6} and \texttt{gpt-4.1} write the most positive narratives averaged across all datasets (0.41 and 0.40), the two small models the least positive (0.29--0.30). Even though Saudi HR applicants have a higher predicted risk (0.64 against 0.48), German Credit narratives are the most negative (0.16 on average) and Saudi HR narratives the most positive (0.49). This difference seems due to the dataset's vocabulary: features and the wording of the predicted outcome count as sentiment words, such as \textit{risk} in German Credit (from ``bad credit risk'', 2.4 times per narrative) or \textit{commitment} and \textit{stability} in Saudi HR. As sentiment is only compared within a dataset, this does not affect our results. 

\begin{table}[!h]
\footnotesize
\caption{Mean net sentiment, $(pos-neg)/(pos+neg)$ based on the Bing Liu opinion lexicon, per generator LLM and dataset under the $PA_{\text{disclosed}}$ condition (range $-1$ to $1$).}
\label{tab:sent_general}
\begin{tabular}{@{}lccccc@{}}
\toprule
Model & German Credit & Saudi HR & Student & Law & All \\
\midrule
\texttt{claude-sonnet-4-6}  & 0.32 & 0.55 & 0.40 & 0.37 & \textbf{0.41} \\
\texttt{gpt-4.1}            & 0.16 & 0.67 & 0.41 & 0.35 & 0.40 \\
\texttt{mistral-large-2512} & 0.17 & 0.39 & 0.33 & 0.35 & 0.31 \\
\texttt{gpt-4o-mini}        & 0.04 & 0.42 & 0.44 & 0.28 & 0.30 \\
\texttt{ministral-14b}      & 0.10 & 0.42 & 0.37 & 0.28 & 0.29 \\
\bottomrule
\end{tabular}
\end{table}

\subsubsection{Differences between social groups}
Six sentiment disparities survive FDR correction, driven mainly by the protected attribute. The common pattern seems to be that when the disclosed PA marks a group commonly associated with higher risk in that domain, the narrative shifts towards a sentiment of risk and failure.

The clearest case is \texttt{gpt-4.1} in Student Performance, where boys receive less positive narratives than girls (0.37 against 0.44; D1, $p_{FDR}=0.011$). The disclosed sex value drives this: the disparity disappears when sex is withheld (D2, $p=0.34$), relabelling girls as boys lowers their sentiment to the level of actual boys (0.44 to 0.36; D4, $p_{FDR}=0.002$), and  girls' and boys' profiles that are both labelled as boys receive the same sentiment (D5, $p=0.74$). Age works the same way: \texttt{gpt-4o-mini} writes more negatively for older German Credit applicants once they are presented as 25 years old (0.04 to $-0.03$; D4, $p_{FDR}=0.048$), and \texttt{mistral-large-2512} writes more negatively for younger students in Student Performance once their age is disclosed (0.42 to 0.29; D3, $p_{FDR}=0.025$). Labelling students' health as bad lowers \texttt{gpt-4o-mini}'s sentiment in Student Performance (0.44 to 0.38; D4, $p_{FDR}=0.005$), but only in narratives that bring up their health (0.50 to 0.39), not in those that do not (0.40 to 0.38).

One disparity stems from feature values instead: \texttt{ministral-14b} writes more negatively for women's German Credit profiles, even when they are labelled as male (0.03 against 0.12 for actual men; D5, $p_{FDR}=0.004$). This difference persists when controlling for predicted risk. Part of it could result from women more often having a negative checking-account balance (48\% against 30\%), which \texttt{ministral-14b} tends to describe harshly (\textit{``your checking account balance was negative, which is a red flag''}), but most of the gap remains unexplained. 

Taken together, sentiment disparities disadvantage groups that are commonly associated with worse outcomes in the domain. Once their protected attribute is disclosed,  boys and students in poor health (in education), and young applicants (in credit scoring) receive a more negative sentiment in their narratives. The exception is \texttt{ministral-14b}, which writes more negatively for women's credit profiles, through their feature values only.

\subsection{Synthesis}
Significant disparities occur in all four datasets, in all three metric families and for all five models. The three metric families all reveal that the LLM's expectations can disregard the input. In faithfulness, expectations about features override the SHAP artefact. In feature coverage and sentiment, expectations about social groups determine whose protected attributes are brought up, how much room is left for secondary features and how the narrative is framed, while feature values determine which secondary features are mentioned. Across all metrics, the disparities mostly disadvantage women (as employees and credit applicants), younger people (as employees, credit applicants and students), boys (as students), non-white law school applicants and students in poor health, though not exclusively.

\section{Discussion}
\label{sec:discussion}

Our results show that LLMs  include their own expectations in the XAINs, about features in faithfulness and about social groups in feature coverage and sentiment. This is in line with earlier observations: \citet{ichmoukhamedov_how_2024} found that LLMs ``correct'' counterintuitive SHAP signs toward intuitive directions, and \citet{silvestri_survey_2025} warn that narratives may favour fluency over faithfulness. These mistakes are not harmless. Because the values that trigger them are unequally distributed, they translate into disparities between social groups. 

We find that withholding the protected attributes, a form of fairness through unawareness \citep{dwork2012fairness}, can remove label-driven feature coverage and sentiment disparities, but not the feature-driven faithfulness disparities, and can even introduce new errors. The direct comparison (D4), which applies the logic of counterfactual fairness \citep{kusner2017counterfactual} to XAINs, makes this distinction visible. Only changing the protected attribute value hardly affects faithfulness, but changes PA and secondary feature mention rates, and the sentiment of the narrative. 

Furthermore, faithfulness follows model cost: the two most expensive models show no faithfulness disparities, as they barely make mistakes in the first place. Fairness in faithfulness in XAINs thus partly becomes something that can be bought, raising an equity concern: institutions with fewer resources, or those that prefer small self-hosted models for data protection, may offer their applicants less faithful explanations. As models improve, faithfulness is likely to keep improving, shifting the focus of fairness research in XAINs toward feature coverage and sentiment, where even the most faithful and expensive models show disparities. The two mitigations are thus complementary: a more faithful model removes the feature-driven faithfulness disparities, and withholding the protected attributes mostly removes the label-driven coverage and sentiment disparities, but neither does both. Only their combination, the two most expensive models with the protected attributes withheld, showed no significant disparity between social groups in this study. This may seem the fairest configuration, but it has limits: it relies on the most expensive models, it sacrifices personalisation, and smaller differences than our sample can detect are not ruled out. It also raises the question of what a fair explanation is in the context of XAINs in the first place, which some findings show.

First, not every significant difference in XAIN content between social groups is unfair by definition as they may reflect genuine differences in the underlying feature values: a narrative that mentions a woman's strong GPA personalises the explanation, whereas a disclaimer triggered by race does not. Distinguishing legitimate personalisation from bias requires further conceptual work, but the decomposition used in this study already draws a clear line. Differences that are driven by the feature values of the profile (D5) can be legitimate, differences driven by the PA label alone (D3, D4) generally are not. Second, equal treatment of groups by the LLM is not sufficient. The plausible falsehoods behind faithfulness disparities are blind to the social group, yet disadvantage some groups more through their feature values distribution. Third, an XAIN has limited length, so any content added to it displaces explanation. A disparity in seemingly harmless content, such as the non-discrimination disclaimers non-white applicants received more often, thus becomes a disparity in how much of their own situation is explained. This concern related to length limits is specific to fairness in XAINs, without analogue in the fields of fairness in XAI or fairness in NLP.

Our findings have implications for auditors, XAIN providers, researchers and regulators. For auditors, fairness in translating the XAI artefact into an XAIN should be audited separately from the predictive model or XAI artefact, as a fair model or artefact can still be translated unfairly. Audits should generate narratives under all three conditions, as the five disparity types follow from them without further generations, and only together can they tell whether differences originate from the PA value, its disclosure or the applicants' profiles. For XAIN providers, LLMs should be chosen for faithfulness, with feature coverage and tone tested separately. Because removing the PAs from the prompt mostly removes label-driven disparities, but leaves the feature-driven faithfulness disparities in place, and can even introduce new errors in less faithful models, withholding the PAs should not be relied upon on its own. For researchers, XAIN evaluation should include counterintuitive feature values and adverse cases without supporting features, and sentiment metrics should be added to existing quality taxonomies \citep{silvestri_survey_2025}. For regulators, the right to a ``clear and meaningful explanation'' under Article 86 of the AI Act \citep{euaiact2024} says nothing about explanation differences across social groups. This is problematic as our results show that the explanations themselves can differ harmfully between groups in a multitude of ways.

Lastly, we state some limitations and future work. We used a single prompt template and single generations at $T=0$, a sentiment lexicon developed for product reviews, and a single extractor validated on 100 narratives. Narratives were generated in English, and social groups were compared as binary splits, without intersectional analysis. We did not study how applicants perceive these narratives: whether the observed differences affect how they understand, trust or contest decisions remains a question for future work, alongside extending the framework to other XAI artefacts, prompts, languages and intersectional groups. 

Finally, we set aside the fairness of the predictive model and the SHAP artefact, which future work could combine with ours to audit the full explanation pipeline. Future work could also research the trade-off between privacy and faithfulness: passing applicant profiles to third-party LLMs poses privacy risks, while the smaller models that can be self-hosted (which avoid this risk) produced less faithful narratives in our study.


\section{Conclusion}
\label{sec:conclusion}
This paper introduced fairness in XAINs as a distinct field of research and proposed a framework that traces disparities between social groups to their source by generating narratives with the protected attributes disclosed, withheld or counterfactually swapped. Applied to four high-risk datasets and five LLMs, it revealed disparities in faithfulness, feature coverage and sentiment, driven either by feature values that trigger plausible falsehoods or by the protected attribute itself. Neither withholding the protected attributes nor using a more expensive, faithful model removes both. As LLMs increasingly turn XAI artefacts into narratives people receive, the translation step must be audited for fairness in its own right.

\section*{Generative AI Usage Statement}

During writing, the authors used Anthropic's Claude Opus 5.5 to check text for grammar, to correct \LaTeX{} code, and to check the consistency of the reported numbers, statistical procedures and descriptions against the code repository\footnote{available at \url{https://github.com/WannesVandenBulck/fairxain}.}. All suggestions were reviewed and revised by the authors. The authors did not use generative AI to conduct literature reviews or to design the study. Generative AI (GitHub Copilot) was only used to assist in writing and debugging the code that implements it, which the authors reviewed and verified. The LLMs used as generators and extractor in the study itself are described in Section~\ref{llm-providers} and are exclusively part of the research, not of the writing process.

\section*{Ethical Considerations Statement}

This study involves no human subjects and collects no new data. All four datasets are publicly available datasets that contain no direct identifiers and that are widely used in fairness research. They were used in accordance with their licences. The main ethical concern is privacy. Generating a narrative requires passing at least part of an applicant's profile to an LLM. In this study, this posed little risk, as all data were already public. In deployment, however, the same pipeline would send sensitive data of real applicants to third-party providers, raising questions of data protection (e.g.\ under the GDPR), data retention and purpose limitation. Self-hosting smaller models avoids this, but our results show that these models produce less faithful narratives, with more disparities between social groups. Institutions may therefore face a trade-off between privacy and explanation quality, which we flag as a risk. Second, comparing social groups requires categorising people, and our binary splits simplify this. Here, race is reduced to white versus non-white, and health and age are split at thresholds. These categories serve the statistical comparison only and are not meant to reflect how individuals identify. Last, our framework is intended to help detect unfair explanations across social groups, but it could also be misused to label narratives as fair based on a limited audit. A lack of significant disparities in our setting does not rule out smaller differences, other prompts or other social groups not included in this study, and should not be read as an endorsement of any model. To support this, we have publicly released all code, prompts, narratives and extractions.

\begin{acks}
Wannes Van den Bulck is supported by a PhD Fellowship Fundamental Research from the Research Foundation Flanders (FWO), grant 11AOT27N. This work was also supported by FWO grant G0G2721N.
\end{acks}

\bibliographystyle{ACM-Reference-Format}
\bibliography{FAIR-XAIN}

\appendix
\section{Statistical Testing Details}
\label{app:stats}

\subsection{General procedures}
\label{app:stats-general}

All tests are two-sided. For each comparison, only narratives for which the metric is non-missing are used (Table~\ref{tab:metrics}). A comparison was not evaluated when fewer than five non-missing observations were available per group (D1, D2, D5) or fewer than five complete pairs (D3, D4), or when the metric showed no variation that could constitute a difference (e.g., identical values in all narratives, or no discordant pairs). Raw $p$-values were corrected using the Benjamini--Hochberg false-discovery-rate procedure ($\alpha = 0.05$), applied separately within batches (Table~\ref{tab:fdr-batches}). Binary and continuous metrics always form separate batches, as they are structurally distinct test families and the continuous metrics are aggregates of the binary ones; unrelated protected attributes, social groups or counterfactual swaps likewise do not share a batch. Effect sizes are odds ratios for binary metrics and rank-biserial correlations for continuous metrics throughout, in the independent, matched or stratified form matching each design.

\begin{table}[h]
\caption{Benjamini--Hochberg correction batches per disparity type.}
\label{tab:fdr-batches}
\footnotesize
\begin{tabular}{@{}ll@{}}
\hline
\textbf{Disparity type} & \textbf{One batch per} \\
\hline
D1, D2 (total, indirect) & dataset $\times$ generator LLM $\times$ protected attribute $\times$ metric type \\
D3 (disclosure) & dataset $\times$ generator LLM $\times$ protected attribute $\times$ social group $\times$ metric type \\
D4, D5 (direct, value) & dataset $\times$ generator LLM $\times$ counterfactual swap $\times$ metric type \\
\hline
\end{tabular}
\end{table}

\subsection{Independent-group comparisons (D1, D2)}
\label{app:independent}

D1 ($PA_{\text{disclosed}}$) and D2 ($PA_{\text{withheld}}$) compare the narratives of the two social groups of a protected attribute (Figure~\ref{fig:datasets}), which consist of different applicants. Binary metrics were compared on the $2 \times 2$ table of group membership against the indicator value, with $a$ and $b$ the counts of 1 and 0 in the first group and $c$ and $d$ those in the second, using Pearson's $\chi^2$ test without continuity correction, or Fisher's exact test when any expected cell count fell below 5. The effect size is the odds ratio $OR = ad/bc$, with 0.5 added to every cell when any cell is zero (Haldane--Anscombe correction). Continuous metrics were compared with the Mann--Whitney $U$ test, with effect size the rank-biserial correlation $r = 2U/(n_1 n_2) - 1$, where $n_1$ and $n_2$ are the group sizes.

\subsection{Paired comparisons (D3, D4)}
\label{app:paired}

D3 and D4 compare two narratives of the \emph{same} instance from the same generator LLM: $PA_{\text{disclosed}}$ versus $PA_{\text{withheld}}$ within one social group (D3), and the counterfactual versus the factual ($PA_{\text{disclosed}}$) narrative within one counterfactual swap (D4). Binary metrics were tested with McNemar's test on the paired $2 \times 2$ table, using the continuity-corrected statistic
\begin{equation}
\chi^2 = \frac{(|b - c| - 1)^2}{b + c},
\end{equation}
where $b$ and $c$ are the counts of the two types of discordant pairs, or its exact binomial form when $b + c < 25$. The effect size is the paired odds ratio $OR = b/c$, with 0.5 added to both counts when either is zero. Continuous metrics were tested with the Wilcoxon signed-rank test on the paired differences, with effect size the matched rank-biserial correlation $r = (n_+ - n_-)/(n_+ + n_-)$, where $n_+$ and $n_-$ are the numbers of positive and negative non-zero differences.

\subsection{Stratified comparisons (D5)}
\label{app:stratified}

D5 compares the counterfactual narratives of a counterfactual swap with the factual ($PA_{\text{disclosed}}$) narratives of instances that actually hold the displayed protected-attribute value, from the same generator LLM. To hold all displayed protected attributes constant, narratives were matched on their displayed protected-attribute profile using coarsened exact matching: the swapped attribute is matched exactly, except for continuous age, which is split at the median; the other protected attributes are matched on the two-group splits of Figure~\ref{fig:datasets}. Each profile forms a stratum, and strata containing only one side were dropped. As both sides consist of different applicants, stratified versions of the tests in Appendix~\ref{app:independent} were used, which reduce to those tests when there is a single stratum. Binary metrics were tested with the Cochran--Mantel--Haenszel test without continuity correction, or, when any stratum had an expected cell count below 5, with the exact stratified test (the exact conditional distribution of the counterfactual count, i.e., the convolution of the per-stratum hypergeometric distributions). The effect size is the Mantel--Haenszel common odds ratio, with 0.5 added to strata containing a zero cell. Continuous metrics were tested with the van Elteren test, a stratified Mann--Whitney $U$ test with stratum weights $1/(N_h + 1)$ and tie correction, with effect size the rank-biserial correlation per stratum, averaged with weights $n_{1h} n_{2h}/N_h$, where $n_{1h}$ and $n_{2h}$ are the sizes of both sides in stratum $h$ and $N_h = n_{1h} + n_{2h}$.

\subsection{Model-level heterogeneity}
\label{app:model-level}

Model comparisons are paired: for a given dataset and condition, every instance has one narrative per model, so each comparison between two models is made on the same instances. Binary indicators were tested with McNemar's test as in Appendix~\ref{app:paired}, with effect size the paired odds ratio $OR = (b + 0.5)/(c + 0.5)$, where $b$ and $c$ are the counts of discordant pairs favouring model 1 and model 2, respectively, so that values above 1 indicate that model 1 wins more of the discordant cases. Continuous metrics were tested with the Wilcoxon signed-rank test, with effect size the paired Cohen's $d_z = \bar{d}/s_d$, where $\bar{d}$ and $s_d$ are the mean and standard deviation of the paired differences.

Comparisons were made at three scopes: within a dataset (overall), within one social group of a dataset (group), and across all four datasets for the continuous metrics (pooled), pairing on the instance within its dataset. Under the $PA_{\text{counterfactual}}$ condition, every instance has one narrative per counterfactual swap. To keep one observation per instance, the continuous metrics were averaged per instance and model over its counterfactual narratives for the overall and pooled scopes, and binary indicators were not tested at these scopes. In place of the group scope, each counterfactual swap was tested separately on all metrics, as all its narratives display the same swapped value.

\clearpage
\section{Tables}
\label{app:tables}
\begin{table}[!h]
\centering
\footnotesize
\setlength{\tabcolsep}{3pt}
\caption{Overview of the LLMs used as generators and extractor, with their batch API prices as listed by the providers at the time of generating and extracting. All narratives were generated on 29--30 October 2026.}
\label{tab:llms_overview}
\begin{tabular}{@{}lllllrr@{}}
\toprule
 & & & & & \multicolumn{2}{c}{Batch price (\$/M tokens)} \\
\cmidrule(l){6-7}
Model & Role & Provider & Weights & Region & Input & Output \\
\midrule
\texttt{claude-sonnet-4-6}            & Generator & Anthropic  & Closed & United States & 1.50  & 7.50 \\
\texttt{gpt-4.1-2025-04-14}           & Generator & OpenAI     & Closed & United States & 1.00  & 4.00 \\
\texttt{mistral-large-2512}           & Generator & Mistral AI & Open   & France        & 0.25  & 0.75 \\
\texttt{ministral-14b-2512}           & Generator & Mistral AI & Open   & France        & 0.10  & 0.10 \\
\texttt{gpt-4o-mini-2024-07-18}       & Generator & OpenAI     & Closed & United States & 0.075 & 0.30 \\
\midrule
\texttt{grok-4.20-0309-non-reasoning} & Extractor & xAI        & Closed & United States & 1.00  & 2.00 \\
\bottomrule
\end{tabular}
\end{table}

\begin{table}[!h]
\caption{Metrics computed per narrative. Each metric is recorded per feature as a binary indicator and, where applicable, aggregated per narrative into a continuous score following Eq.~(\ref{eq:xa}). Indicators for which nothing could be extracted from the narrative are treated as missing rather than incorrect. Secondary features are all features outside the top-3 SHAP features, excluding the protected attributes.}
\label{tab:metrics}
\footnotesize
\setlength{\tabcolsep}{2.6pt}
\begin{tabular}{@{}>{\raggedright\arraybackslash}p{2.3cm}>{\raggedright\arraybackslash}p{1.6cm}>{\raggedright\arraybackslash}p{3.3cm}>{\raggedright\arraybackslash}p{2.9cm}>{\raggedright\arraybackslash}p{3.6cm}@{}}
\hline
\textbf{Metric} & \textbf{Type} & \textbf{Per feature (binary)} & \textbf{Per narrative (continuous)} & \textbf{Definition} \\
\hline
Rank Accuracy (RA) & Faithfulness & \texttt{rank\_n\_scoring} & \texttt{rank\_position\_\allowbreak accuracy} & The feature presented at rank $n$ is the feature at SHAP rank $n$. \\
Sign Accuracy (SA) & Faithfulness & \texttt{rank\_n\_sign\_scoring} & \texttt{sign\_accuracy} & The stated direction of influence of the SHAP rank-$n$ feature is correct, wherever the narrative places it among its three most important features. \\
Value Accuracy (VA) & Faithfulness & \texttt{rank\_n\_value\_scoring}, \texttt{<PA>\_scoring}, \texttt{other\_<feature>\_\allowbreak value\_scoring} & \texttt{value\_accuracy} & The stated value of a top-3 feature, protected attribute or secondary feature exactly matches the instance's feature value. \\
Predicted-probability accuracy & Faithfulness & \texttt{predicted\_\allowbreak probability\_scoring} & -- & The stated predicted probability lies within 0.01 of the model's prediction. \\
Completeness & Coverage & \texttt{rank\_n\_mentioned} & -- & The rank-$n$ SHAP feature is mentioned. \\
PA mention & Coverage & \texttt{<PA>\_mentioned} & \texttt{prop\_pa\_\allowbreak mentioned} & The protected attribute is mentioned; per narrative, the share of protected attributes mentioned. \\
Secondary-feature mention & Coverage & \texttt{other\_<feature>\_\allowbreak mentioned} & \texttt{prop\_other\_\allowbreak mentioned} & The secondary feature is mentioned; per narrative, the share of secondary features mentioned. \\
\hline
\end{tabular}
\end{table}

\begin{table*}[!h]
\scriptsize
\setlength{\tabcolsep}{3pt}
\caption{All 19 FDR-significant faithfulness disparities. \textit{Groups}: for D1 and D2, the two social groups compared; for D3, the narratives of one group with the protected attributes disclosed vs withheld; for D4, narratives with the swapped protected attribute vs the factual narratives of the same instances; for D5, narratives of one group whose protected attribute was swapped vs narratives of the group that actually holds these PA values. \textit{Values}: the metric for the first vs the second side of the comparison (\%). RA, SA, VA: rank, sign and value accuracy, per narrative or for a single rank.}
\label{tab:faith_findings}
\begin{tabular}{@{}llllllll@{}}
\toprule
 & Dataset & Model & Groups & Metric & Values & $p_{FDR}$ & Mechanism \\
\midrule
D1 & German Credit & \texttt{gpt-4o-mini} & women vs men & RA (rank 1) & 65.0 vs 90.0 & 0.040 & prior: status first \\
D1 & Saudi HR & \texttt{gpt-4o-mini} & 21--30 vs 31+ & SA (rank 1) & 34.1 vs 67.3 & 0.024 & prior: experience \\
D1 & Saudi HR & \texttt{gpt-4o-mini} & women vs men & SA & 56.3 vs 72.7 & 0.022 & prior: experience \\
D1 & Saudi HR & \texttt{gpt-4o-mini} & women vs men & SA (rank 1) & 36.5 vs 70.2 & 0.019 & prior: experience \\
\midrule
D2 & German Credit & \texttt{gpt-4o-mini} & $\geq$30 vs $<$30 & RA & 63.7 vs 85.6 & 0.010 & prior: status first \\
D2 & German Credit & \texttt{gpt-4o-mini} & $\geq$30 vs $<$30 & RA (rank 3) & 57.1 vs 86.4 & 0.029 & prior: status first \\
D2 & Student & \texttt{ministral-14b} & good vs poor health & SA & 74.5 vs 55.7 & 0.045 & justification \\
D2 & Student & \texttt{mistral-large-2512} & good vs poor health & SA & 92.4 vs 82.9 & 0.044 & justification \\
\midrule
D3 & Saudi HR & \texttt{gpt-4o-mini} & 21--30: disclosed vs withheld & VA & 94.8 vs 81.5 & 0.025 & value fabrication \\
D3 & Saudi HR & \texttt{gpt-4o-mini} & no health issues: disclosed vs withheld & VA (rank 2) & 91.4 vs 71.4 & 0.012 & value fabrication \\
D3 & Saudi HR & \texttt{gpt-4o-mini} & women: disclosed vs withheld & RA & 89.9 vs 79.2 & 0.021 & unclear \\
D3 & Saudi HR & \texttt{gpt-4o-mini} & women: disclosed vs withheld & VA & 95.6 vs 84.2 & 0.010 & value fabrication \\
D3 & Saudi HR & \texttt{ministral-14b} & 21--30: disclosed vs withheld & SA & 69.3 vs 80.7 & 0.046 & prior: experience \\
D3 & Saudi HR & \texttt{ministral-14b} & 31+: disclosed vs withheld & VA & 90.6 vs 96.7 & 0.011 & unclear \\
\midrule
D4 & German Credit & \texttt{mistral-large-2512} & $\geq$30 $\to$ 25: swapped vs factual & RA & 82.7 vs 72.0 & 0.042 & unclear \\
\midrule
D5 & Saudi HR & \texttt{gpt-4o-mini} & 21--30 $\to$ 31+: swapped vs actual & SA (rank 1) & 31.1 vs 68.2 & 0.026 & prior: experience \\
D5 & Saudi HR & \texttt{ministral-14b} & 31+ $\to$ 21--30: swapped vs actual & VA & 87.6 vs 96.0 & 0.045 & unclear \\
D5 & Saudi HR & \texttt{ministral-14b} & women $\to$ men: swapped vs actual & RA (rank 3) & 90.6 vs 63.8 & 0.021 & unclear \\
D5 & Saudi HR & \texttt{mistral-large-2512} & women $\to$ men: swapped vs actual & RA & 96.2 vs 83.0 & 0.013 & unclear \\
\bottomrule
\end{tabular}
\end{table*}

\begin{table*}[!h]
\scriptsize
\setlength{\tabcolsep}{3pt}
\caption{All 34 FDR-significant feature-coverage disparities. \textit{Groups}: for D1 and D2, the two social groups compared; for D3, the narratives of one group with the protected attributes disclosed vs withheld; for D4, narratives with the swapped protected attribute vs the factual narratives of the same instances; for D5, narratives of one group whose protected attribute was swapped vs narratives of the group that actually holds these PA values. \textit{Metric}: whether a feature is mentioned, or the per-narrative proportion of protected attributes (PA prop.) or secondary features (Sec.\ prop.) mentioned. \textit{Values}: the metric for the first vs the second side of the comparison (\%). \textit{Mechanism}: (i) the protected attribute enters the narrative, (ii) feature values.}
\label{tab:cov_findings}
\begin{tabular}{@{}llllllll@{}}
\toprule
 & Dataset & Model & Groups & Metric & Values & $p_{FDR}$ & Mechanism \\
\midrule
D1 & Law & \texttt{claude-sonnet-4-6} & non-white vs white & PA prop. & 38.0 vs 12.0 & 0.017 & (i) disclaimer \\
D1 & Law & \texttt{claude-sonnet-4-6} & non-white vs white & gender mentioned & 38.0 vs 12.0 & 0.016 & (i) disclaimer \\
D1 & Law & \texttt{claude-sonnet-4-6} & non-white vs white & race mentioned & 38.0 vs 12.0 & 0.016 & (i) disclaimer \\
D1 & Student & \texttt{gpt-4o-mini} & good vs poor health & PA prop. & 0.6 vs 9.3 & $<$0.001 & (i) health \\
D1 & Student & \texttt{gpt-4o-mini} & good vs poor health & health mentioned & 1.8 vs 27.9 & 0.003 & (i) health \\
\midrule
D2 & German Credit & \texttt{gpt-4o-mini} & $\geq$30 vs $<$30 & Sec.\ prop. & 3.1 vs 1.0 & 0.042 & (ii) feature values \\
D2 & Law & \texttt{ministral-14b} & non-white vs white & LSAT mentioned & 66.7 vs 11.1 & 0.017 & unclear \\
\midrule
D3 & Law & \texttt{claude-sonnet-4-6} & men: disclosed vs withheld & PA prop. & 22.0 vs 2.0 & 0.019 & (i) disclaimer \\
D3 & Law & \texttt{claude-sonnet-4-6} & men: disclosed vs withheld & gender mentioned & 22.0 vs 2.0 & 0.041 & (i) disclaimer \\
D3 & Law & \texttt{claude-sonnet-4-6} & men: disclosed vs withheld & race mentioned & 22.0 vs 2.0 & 0.041 & (i) disclaimer \\
D3 & Law & \texttt{claude-sonnet-4-6} & non-white: disclosed vs withheld & PA prop. & 38.0 vs 8.0 & 0.001 & (i) disclaimer \\
D3 & Law & \texttt{claude-sonnet-4-6} & non-white: disclosed vs withheld & gender mentioned & 38.0 vs 8.0 & 0.002 & (i) disclaimer \\
D3 & Law & \texttt{claude-sonnet-4-6} & non-white: disclosed vs withheld & race mentioned & 38.0 vs 8.0 & 0.002 & (i) disclaimer \\
D3 & Law & \texttt{claude-sonnet-4-6} & non-white: disclosed vs withheld & Sec.\ prop. & 14.7 vs 28.0 & 0.025 & (i) crowding out \\
D3 & Law & \texttt{claude-sonnet-4-6} & women: disclosed vs withheld & PA prop. & 28.0 vs 6.0 & 0.005 & (i) disclaimer \\
D3 & Law & \texttt{claude-sonnet-4-6} & women: disclosed vs withheld & gender mentioned & 28.0 vs 6.0 & 0.005 & (i) disclaimer \\
D3 & Law & \texttt{claude-sonnet-4-6} & women: disclosed vs withheld & race mentioned & 28.0 vs 6.0 & 0.005 & (i) disclaimer \\
D3 & Saudi HR & \texttt{gpt-4o-mini} & 31+: disclosed vs withheld & Sec.\ prop. & 3.1 vs 5.1 & 0.036 & unclear \\
D3 & Student & \texttt{gpt-4o-mini} & $\geq$17: disclosed vs withheld & PA prop. & 4.2 vs 0.0 & 0.049 & (i) health \\
D3 & Student & \texttt{gpt-4o-mini} & poor health: disclosed vs withheld & PA prop. & 9.3 vs 0.0 & 0.003 & (i) health \\
D3 & Student & \texttt{gpt-4o-mini} & poor health: disclosed vs withheld & health mentioned & 27.9 vs 0.0 & 0.010 & (i) health \\
D3 & Student & \texttt{gpt-4o-mini} & women: disclosed vs withheld & PA prop. & 5.6 vs 0.0 & 0.009 & (i) health \\
D3 & Student & \texttt{gpt-4o-mini} & women: disclosed vs withheld & health mentioned & 16.7 vs 0.0 & 0.047 & (i) health \\
D3 & Student & \texttt{ministral-14b} & poor health: disclosed vs withheld & PA prop. & 5.4 vs 0.0 & 0.049 & (i) health \\
\midrule
D4 & Law & \texttt{claude-sonnet-4-6} & non-white $\to$ white: swapped vs factual & PA prop. & 10.0 vs 38.0 & 0.007 & (i) disclaimer \\
D4 & Law & \texttt{claude-sonnet-4-6} & non-white $\to$ white: swapped vs factual & gender mentioned & 10.0 vs 38.0 & 0.012 & (i) disclaimer \\
D4 & Law & \texttt{claude-sonnet-4-6} & non-white $\to$ white: swapped vs factual & race mentioned & 10.0 vs 38.0 & 0.012 & (i) disclaimer \\
D4 & Student & \texttt{gpt-4o-mini} & good $\to$ bad health: swapped vs factual & PA prop. & 14.6 vs 0.6 & $<$0.001 & (i) health \\
D4 & Student & \texttt{gpt-4o-mini} & good $\to$ bad health: swapped vs factual & health mentioned & 43.9 vs 1.8 & $<$0.001 & (i) health \\
D4 & Student & \texttt{gpt-4o-mini} & poor $\to$ very good health: swapped vs factual & PA prop. & 0.0 vs 9.3 & 0.003 & (i) health \\
D4 & Student & \texttt{gpt-4o-mini} & poor $\to$ very good health: swapped vs factual & health mentioned & 0.0 vs 27.9 & 0.011 & (i) health \\
\midrule
D5 & Law & \texttt{claude-sonnet-4-6} & men $\to$ women: swapped vs actual & GPA mentioned & 12.9 vs 53.6 & 0.022 & (ii) feature values \\
D5 & Law & \texttt{gpt-4o-mini} & men $\to$ women: swapped vs actual & GPA mentioned & 3.2 vs 42.9 & 0.013 & (ii) feature values \\
D5 & Saudi HR & \texttt{mistral-large-2512} & health issues $\to$ none: swapped vs actual & Job\_Support mentioned & 42.9 vs 7.0 & 0.029 & (ii) feature values \\
\bottomrule
\end{tabular}
\end{table*}

\begin{table*}[!h]
\scriptsize
\setlength{\tabcolsep}{3pt}
\caption{All 6 FDR-significant sentiment disparities. \textit{Groups}: for D1, the two social groups compared; for D3, the narratives of one group with the protected attributes disclosed vs withheld; for D4, narratives with the swapped protected attribute vs the factual narratives of the same instances; for D5, narratives of one group whose protected attribute was swapped vs narratives of the group that actually holds these PA values. \textit{Values}: mean net sentiment for the first vs the second side of the comparison (range $-1$ to $1$). \textit{Attribute named}: whether the narratives mention the protected attribute that shifts the tone.}
\label{tab:sent_findings}
\begin{tabular}{@{}llllllll@{}}
\toprule
 & Dataset & Model & Groups & Values & $p_{FDR}$ & Driver & Attribute named \\
\midrule
D1 & Student & \texttt{gpt-4.1} & girls vs boys & 0.44 vs 0.37 & 0.011 & label & no \\
\midrule
D3 & Student & \texttt{mistral-large-2512} & $<$17: disclosed vs withheld & 0.29 vs 0.42 & 0.025 & label & no \\
\midrule
D4 & German Credit & \texttt{gpt-4o-mini} & $\geq$30 $\to$ 25: swapped vs factual & $-$0.03 vs 0.04 & 0.048 & label & no \\
D4 & Student & \texttt{gpt-4.1} & girls $\to$ boys: swapped vs factual & 0.36 vs 0.44 & 0.002 & label & no \\
D4 & Student & \texttt{gpt-4o-mini} & good $\to$ bad health: swapped vs factual & 0.38 vs 0.44 & 0.005 & label & yes \\
\midrule
D5 & German Credit & \texttt{ministral-14b} & women $\to$ men: swapped vs actual & 0.03 vs 0.12 & 0.004 & feature values & no \\
\bottomrule
\end{tabular}
\end{table*}

\clearpage
\section{Figures}
\label{app:figures}

\begin{figure}[!h]
  \centering
  \includegraphics[width=0.8\linewidth]{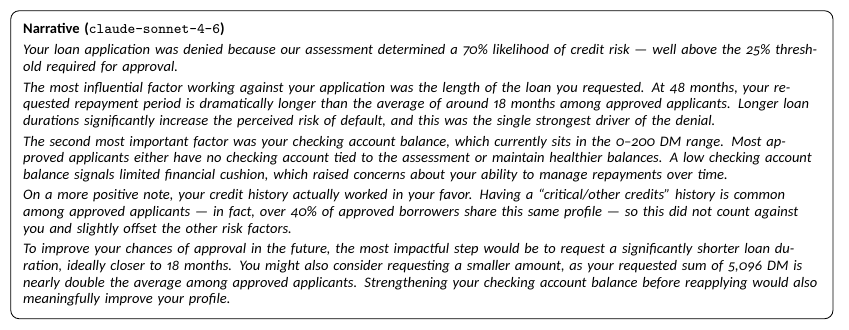}
  \caption{Full narrative generated by \texttt{claude-sonnet-4-6} for German Credit instance 18 under the $PA_{\text{disclosed}}$ condition, of which part is shown in Fig.~\ref{fig:narrativeintro}. This narrative is fully faithful (value, sign and rank accuracy of 100\%, correct predicted probability), mentions all top-3 SHAP features, no protected attributes and one secondary feature (the credit amount), and has a net sentiment of 0.20.}
  \label{fig:fullnarrative}
  \Description{An example of a full narrative}
\end{figure}

\begin{figure}[!h]
  \centering
  \includegraphics[width=0.6\linewidth]{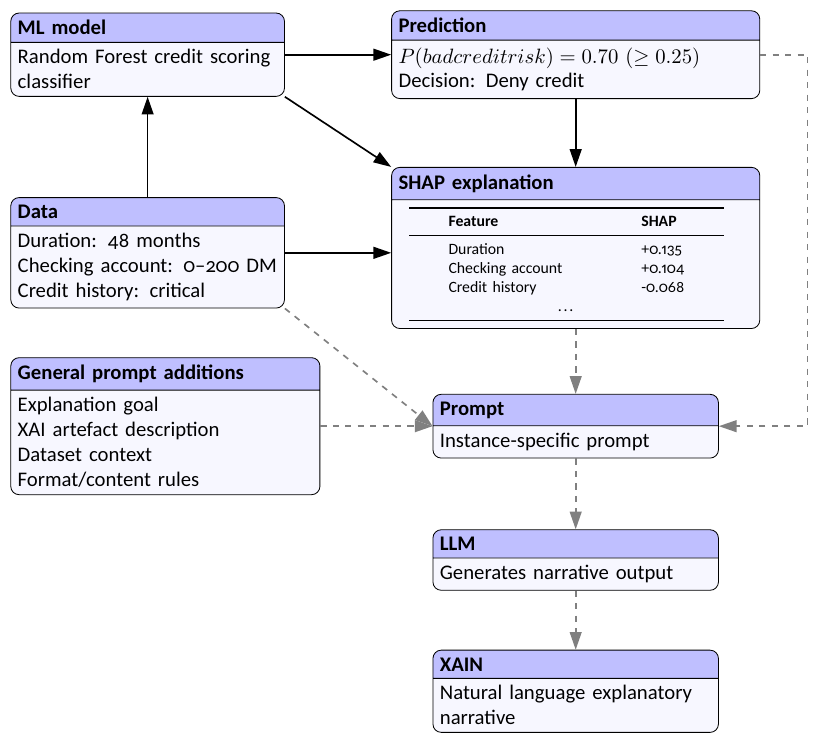}
  \caption{Unified pipeline for generating Explainable AI Narratives (XAINs), illustrated for German Credit instance 18. The grey, dotted arrows depict the extra steps in the explanation pipeline unique to XAINs. The figure is the authors' own work.}
  \label{fig:XAINpipeline}
  \Description{Unified pipeline for generating Explainable AI Narratives (XAINs), illustrated for a credit-scoring example. The grey, dotted arrows depict the extra steps in the explanation pipeline unique to XAINs}
\end{figure}

\begin{figure}[!h]
\centering
\includegraphics[width=\linewidth, height=0.9\textheight, keepaspectratio]{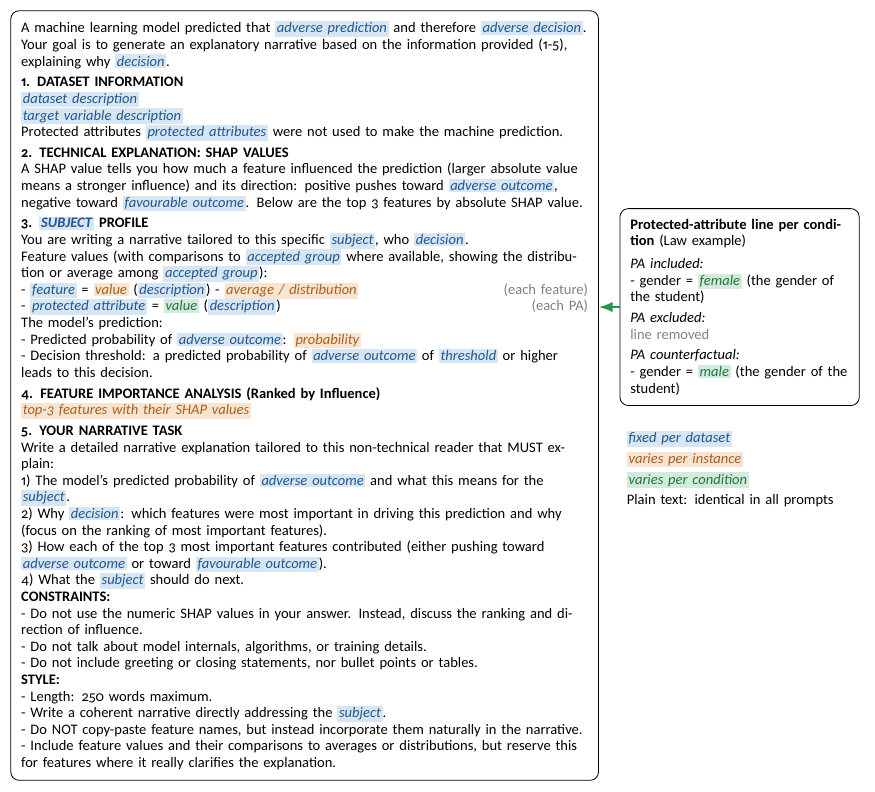}
\caption{Template of the generation prompt. Plain text is identical in all prompts. Highlighted parts are filled in per dataset (blue), per instance (orange) or per generation condition (green). The right panel shows the protected-attribute line under the three conditions, using the Law dataset as an example.}
\label{fig:generator_prompt}
  \Description{Template of the generation prompt.}
\end{figure}

\begin{figure}[!h]
\centering
\includegraphics[width=\linewidth]{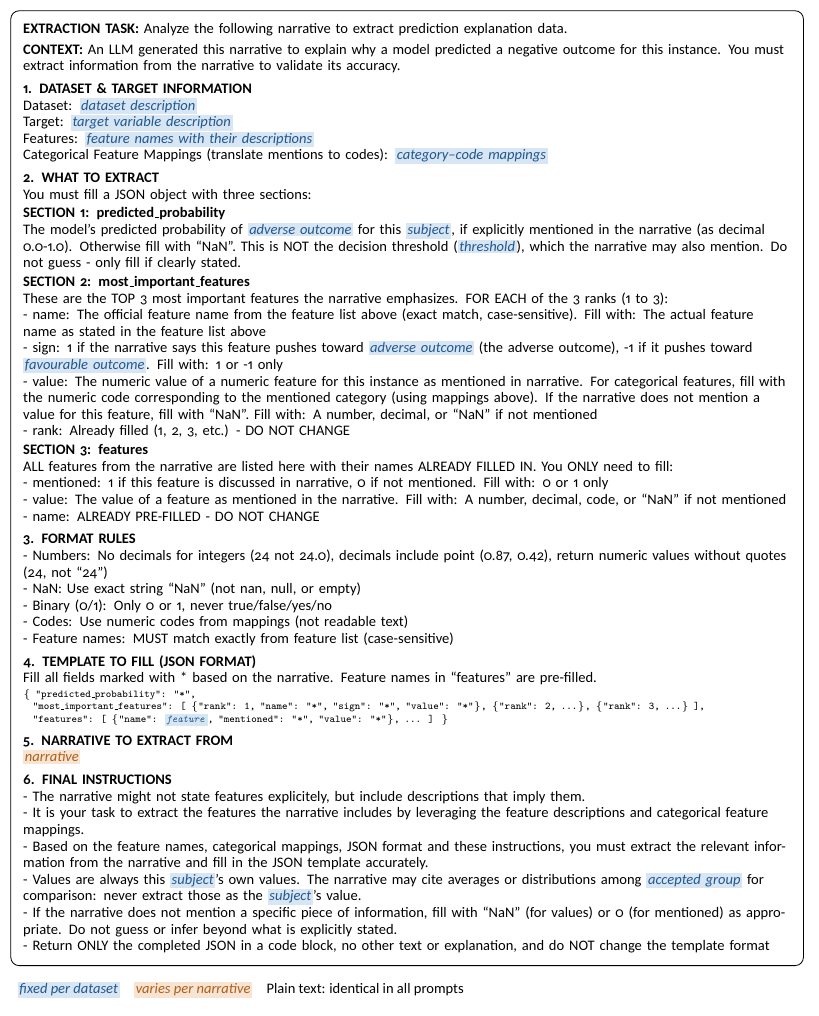}
\caption{Template of the extraction prompt. Plain text is identical in all prompts. Highlighted parts are filled in per dataset (blue) or per narrative (orange). The JSON template is cut off: in the prompts, it lists all features of the dataset.}
\label{fig:extractionprompt}
  \Description{Example extraction prompt for instance 0 from the Law dataset (part 1 of 2).}
\end{figure}

\end{document}